\documentclass{article}
\usepackage[utf8]{inputenc}
\usepackage{authblk}
\usepackage{setspace}
\usepackage[margin=1.25in]{geometry}
\usepackage{graphicx}
\usepackage{subcaption}
\usepackage{amsmath}
\usepackage{lineno}
\usepackage{hyperref}
\hypersetup{
    colorlinks=true,    
    linkcolor=blue,     
    citecolor=blue,     
}
\usepackage{booktabs}
\usepackage{array}
\usepackage{multirow}
\usepackage{textgreek}

\usepackage[style=nejm, 
citestyle=numeric-comp,
sorting=none]{biblatex}
\title{Nigerian Schizophrenia EEG Dataset (NSzED) Towards Data-Driven Psychiatry in Africa.}

\author[1*$\dag$]{E.O. Olateju}
\author[1$\dag$]{K.P. Ayodele}
\author[1]{S.K. Mosaku}

\affil[1,2]{Department of electronic and Electrical Engineering, Obafemi Awolowo University, Ile-Ife, Nigeria.}
\affil[3]{College of Health Sciences, Obafemi Awolowo University, Ile-Ife, Nigeria.}
\affil[1,2]{eoolateju@student.oauife.edu,ng, kpayodele@oauife.edu.ng}
\affil[$\dag$]{These authors contributed equally to this work.}

\date{}

\begin{document}

\maketitle

\begin{abstract}
This work has been carried out to improve the dearth of high-quality EEG datasets used for schizophrenia diagnostic tools development and studies from populations of developing and underdeveloped regions of the world. To this aim, the presented dataset contains international 10/20 system EEG recordings from West African subjects of Nigerian origin in restful states, mental arithmetic task execution states and while passively reacting to auditory stimuli, the first of its kind from the region and continent.

The subjects are divided into patients and healthy controls and recorded from 37 patients and 22 healthy control subjects identified by the Mini International Schizophrenia Interview (MINI) and also assessed by the Positive and Negative Symptoms Scale (PANSS) and the World Health Organization Disability Assessment Schedule (WHODAS). All patients are admitted schizophrenia patients of the Mental Health Ward, Medical Outpatient Department of the Obafemi Awolowo University Teaching Hospital Complex (OAUTHC, Ile-Ife) and its subsidiary Wesley Guild Hospital Unit (OAUTHC, Ilesa). Controls are drawn from students and clinicians who volunteered to participate in the study at the Mental Health Ward of OAUTHC and the Wesley Guild Hospital Unit. 
This dataset is the first version of the Nigerian schizophrenia dataset (NSzED) and can be used by the neuroscience and computational psychiatry research community studying the diagnosis and prognosis of schizophrenia using the electroencephalogram signal modality.
\end{abstract}


\section{Introduction}

Mental disorders encompass a wide range of conditions affecting cognition and behavior, with varying severity. Common worldwide, they affect millions of individuals. Schizophrenia, a chronic mental health disorder significantly influences cognition, emotion, and behavior earning it the moniker ”heartland of psychiatry” due to its pervasive symptoms \cite{goodwin2007heartland}. Those with schizophrenia may struggle to discern reality, experiencing hallucinations, delusions, and disorganized speech. While disabling, effective treatment enables many to lead full lives \cite{tandon2009schizophrenia}.

Affecting approximately 1\% of the global population, schizophrenia typically emerges in early adulthood (18-25 years), with equal prevalence in men and women. Diagnosis relies on psychiatric evaluation, medical record review, and sometimes brain imaging to rule out other conditions \cite{jablensky2010diagnostic}. The disorder manifests through positive, negative and cognitive  symptoms.

Diagnosing mental disorders, including schizophrenia, traditionally relies on subjective clinical evaluation and observational criteria, leading to inherent inconsistencies \cite{jablensky2010diagnostic}. Challenges in accurate diagnosis stem from the complex and heterogeneous symptomatology of schizophrenia, often overlapping with other mental conditions \cite{patel2014schizophrenia,barros2021advanced}. Despite these challenges, empirical studies reveal cognitive impairments, driving a trend in computational psychiatry to identify objective neurophysiological markers for systematic schizophrenia diagnosis in computational psychiatry \cite{perrottelli2022unveiling}.

Numerous neuroimaging tools explore neural functioning in schizophrenia, with electroencephalography (EEG) standing out for its ability to investigate neural underpinnings and functional anomalies. Despite its lower spatial resolution, its non-invasive nature and high temporal resolution make it well-suited for analyzing neural behaviors, and other time-sensitive markers relevant in schizophrenia studies \cite{crouch2018detection, yen2023exploring}. Several existing EEG schizophrenia studies show impressive performance in early-stage identification and remission prediction. However, these datasets have limitations. Commonly, small sample sizes from specific social or cultural regions can impact statistical power and limit conclusive insights. Additionally, many datasets may be biased toward certain ethnic or cultural groups, affecting generalizability and global adoption of resulting models. Variability in EEG recording protocols, equipment, and environmental conditions introduces challenges in standardization. Longitudinal data arse crucial for studying schizophrenia progression, but many datasets lack sufficient information, making it difficult to study the evolution of the disorder. Addressing the issue of limited diversity is essential, thus prompting the creation of the new schizophrenia dataset NSzED (Nigeria’s Schizophrenia Dataset).

A challenge in developing EEG-based diagnostic tools for schizophrenia is the limited availability of geographically diverse datasets. Current research predominantly originates from developed regions, leaving populations from developing regions underrepresented \cite{soria2023method, perrottelli2022unveiling, barros2021advanced, ko2022eeg}. To address this, we present NSzED , a novel collection of EEG data from Nigeria, West Africa, encompassing both healthy controls and clinically diagnosed schizophrenia patients. NSzED facilitates the evaluation and refinement of existing EEG diagnostic models and the potential discovery of novel markers for underrepresented groups.

NSzED is part of a broader initiative with the acronym G.E.N.E.R.I.S to develop an EEG feature framework that adequately addresses the heterogeneity of schizophrenia. A significant obstacle in EEG modeling for schizophrenia is mapping neural signal anomalies to the broad spectrum of the disease manifestations \cite{barros2021advanced}. As such, NSzED incorporates two EEG data collection schemes designed for multimodal feature extractions, schizophrenia markers, and parallel analysis of these markers.

After an extensive literature review highlighting event-related potentials (ERPs), event-related oscillatory dynamics, and neurophysiological complexity metrics as primary EEG-derived markers for schizophrenia \cite{barros2021advanced}, the NSzED protocol integrates task paradigms tailored for computing key biomarkers such as mismatch negativity (MMN), steady state responses (SSR), and entropy-based measures.

With diverse biomarkers for each subject, sophisticated feature fusion techniques enable comparisons to assess the diagnostic efficacy of different EEG metrics \cite{agarwal2023fusion, goshvarpour2020schizophrenia} (see \hyperlink{tab:DAQ_protocol}{Table 1}). In essence, NSzED provides a structured framework for mapping schizophrenia’s neural signatures in a multidimensional space, facilitating advanced computational analyses. This paper details the rationale, design criteria, data collection methods, and organization of NSzED.

\begin{table}
    \centering
    \caption{Data-Acquisition Protocol and Biomarkers Associated With Each Phase of Acquisition}
    \begin{tabular}{>{}m{2cm}>{}m{2cm}>{}m{4cm}>{}m{4cm}}
        \toprule
        \small \textbf{Phase} & \small \textbf{Scheme One} & \small \textbf{Scheme Two} & \small \textbf{Biomarkers (Scheme One/Scheme Two)}\\
        \midrule
        \small Phase One & \small Rest & \small Rest & \small Rest State Features \\
        \addlinespace 
        \small Phase Two & \small Arithmetic Task & \small Arithmetic Task & \small Complexity Measures \\
        \addlinespace
        \small Phase Three & \small Rest & \small Fixed Frequency Auditory Stimuli & \small Rest State Features/Steady State Responses \\
        \addlinespace
        \small Phase Four & \small Auditory Oddball Task & \small Rest & \small Mismatch Negativity/Rest State Features \\
        \addlinespace
        \small Phase Five & \small - & \small Auditory Oddball Task & \small Mismatch Negativity \\
        \bottomrule
    \end{tabular}
    \label{tab:DAQ_protocol}
    \hypertarget{tab:DAQ_protocol}{}
\end{table}

\section{NSzED Overview}
\subsection{EEG Recording}
The NSzED acquisition uses the international 10/20 electrode system for EEG data acquisition, covering symmetric anterior frontal (Fp1, Fp2), frontal (F3, F4, F7, F8), central (C3, C4, Cz), temporal (T3, T4, T5, T6), parietal (P3, P4, Pz), and occipital areas (O1, O2).

EEG data is collected with two machines: contek-2400 (200 Hz), and BrainAtlas Discovery-24E (256 Hz). These different sampling rates capture brain activity at varying temporal resolutions, enabling comprehensive data analysis.

\subsection{EEG Recording Protocol}
The NSzED data acquisition for contek-2400 made use of scheme one while that of the BrainAtlas Discovery-24E made use of scheme two of the acquisition protocols which are both designed for computing various EEG schizophrenia features or biomarkers. The schemes included tasks such as an auditory (oddball) stimulus passive attention task making use of a 1KHz, 100ms standard tone, 1KHz, 200ms duration deviant and a 3KHz, 100ms frequency deviant in the oddball experiment. Also included is 40Hz fixed frequency auditory stimulus passive attention task for discovery-24E recordings, resting recording, and mental arithmetic task execution. The schemes are described in \hyperlink{tab:DAQ_protocol}{Table 1}.

Auditory stimuli were presented through disposable earphones, and arithmetic task instructions were conveyed visually and verbally in the subjects' native language, based on two synchronization schemes. The first scheme used visual and verbal cues (recorded audio) in the subjects' native language, with a fixed time recording starting from the onset of the first arithmetic problem. The second scheme offered flexibility, delivering instructions via recorded audio and arithmetic problems through speech from clinicians, both using disposable earphones. In the second scheme, a hand-held clicker with five buttons, including two used during the arithmetic phase, was employed. Clinicians pressed one button to signal the onset of an arithmetic problem and the second to mark the subject's response within a preset recording time. The first scheme was exclusive to contek-2400 recordings, while the second was used with discovery-24E recordings. \hyperlink{tab:cue_synchronization}{Table 2} describes the cue and instruction delivery mechanism adopted for each phase, recording protocol and arithmetic task synchronization scheme.

This standardized protocol aimed to capture diverse brain responses associated with schizophrenia and enable the computation of multiple EEG-based biomarkers.

\begin{table}
    \centering
    \caption{Cue and Instructions Delivery Methods}
    \begin{tabular}{>{}m{2cm}>{}m{5cm}>{}m{5cm}}
        \toprule
        \small \textbf{Phase} & \small \textbf{Scheme (Protocol) One} & \small \textbf{Scheme (Protocol) Two} \\
        \midrule
        \small Phase One & \small Verbal \& Visual Instructions & \small Verbal \& Visual Instructions \\
        \addlinespace 
        \small Phase Two & \small Synchronization scheme one & \small Synchronization scheme two \\
        \addlinespace
        \small Phase Three & \small Verbal \& Visual Instructions & \small Verbal Instructions \\
        \addlinespace
        \small Phase Four & \small Verbal \& Visual Instructions & \small Verbal \& Visual Instructions \\
        \addlinespace
        \small Phase Five & \small - & \small Verbal \& Visual Instructions \\
        \bottomrule
    \end{tabular}
    \label{tab:cue_synchronization}
    \hypertarget{tab:cue_synchronization}{}
\end{table}

\subsection{EEG Records Selection}
The EEG recording quality was verified by two specialized clinicians, including an electrophysiologist, leading to the exclusion of a specific number of low-quality recordings from NSzED. Additionally, two recordings were excluded—one due to the subject’s history of substance use and the other due to an active epileptic condition.

\subsection{Participants}
Participants were selected from mental health-related units at Obafemi Awolowo University Teaching Hospital College (OAUTHC), Ile-Ife, Nigeria (see \hyperlink{fig:demographic_plots}{Figure 1} and \hyperlink{fig:metric_plots}{Figure 2} for statistical plots of subjects). The patients are made up of 20 females and 17 males, while the healthy controls are made up of 17 females and 5 males. The Yoruba ethnic group is over represented due to the geographical location of data collection within the country (refer to \hyperlink{fig:demographic_plots}{Figure 1} for the Histogram of Subjects' Preferred Language). Approximately 75\% of subjects are are above 25 years of age. All subjects underwent assessment using the Mini International Schizophrenia Interview (MINI), Positive and Negative Symptoms Scale (PANSS), and World Health Organization Disability Assessment Schedule (WHODAS).

\begin{figure}
    \centering
    \includegraphics[width=1\textwidth]{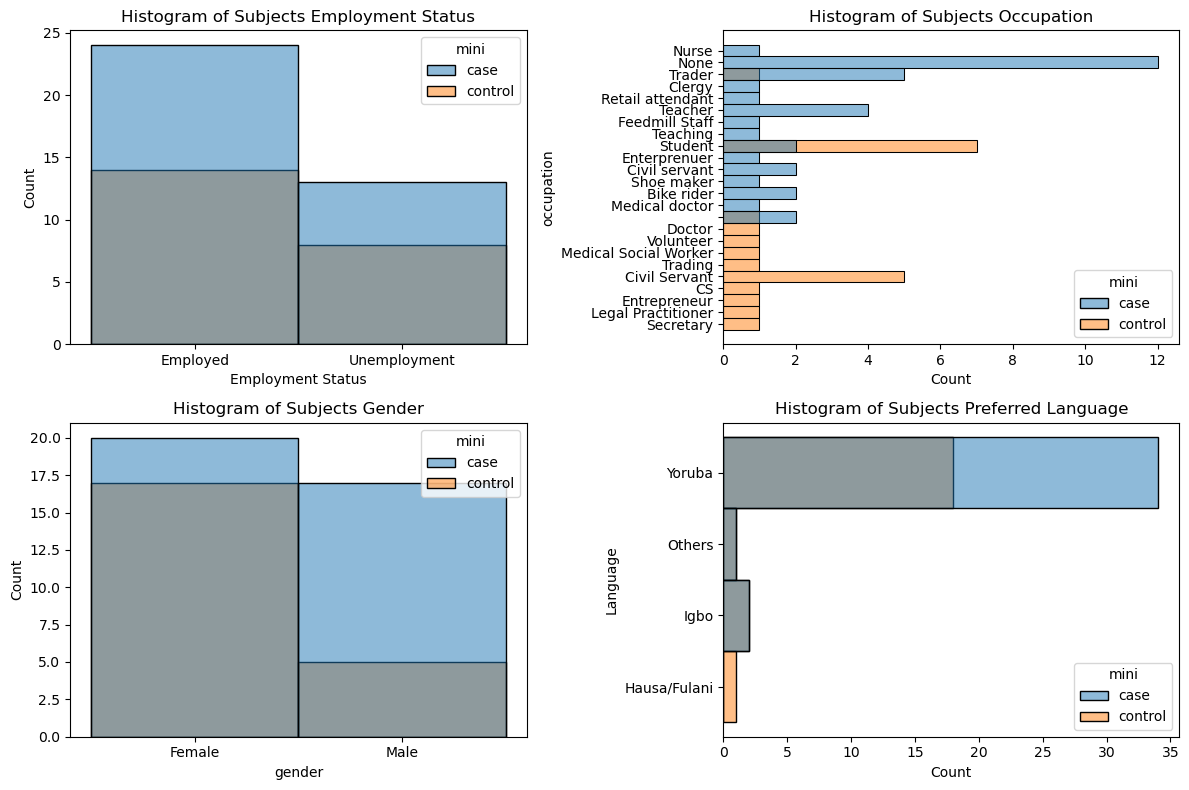}
    \caption{Demographic Information Plots}
    \label{fig:demographic_plots}
    \hypertarget{fig:demographic_plots}{}
\end{figure}

\subsection{Dataset Organization and Format Description}
All NSzED data are available in the NSzED database, organized into release versions. This paper introduces the first version, NSzED-v1. Within each NSzED version folder in the database, there are subfolders for recordings conducted with contek and discovery-24E devices. These device folders further categorize recordings by subjects as folders, with sub-folders for each recording session of the subject. Session folders include recordings ranging from four to five phases based on the recording device and protocol, stored in EDF (European Data Format), along with a GNR file detailing the adopted protocol, arithmetic task synchronization scheme, and recording device. Additionally, each device folder features a .sav file containing preprocessed EEG data in a format suitable for direct use in feature computation. Accompanying spreadsheets in each device folder provide clinical and demographic data. Finally, each subject folder contains a GNR (Global Name Registry) file with subject-specific metadata.

\begin{figure}
    \centering
    \includegraphics[width=1\textwidth]{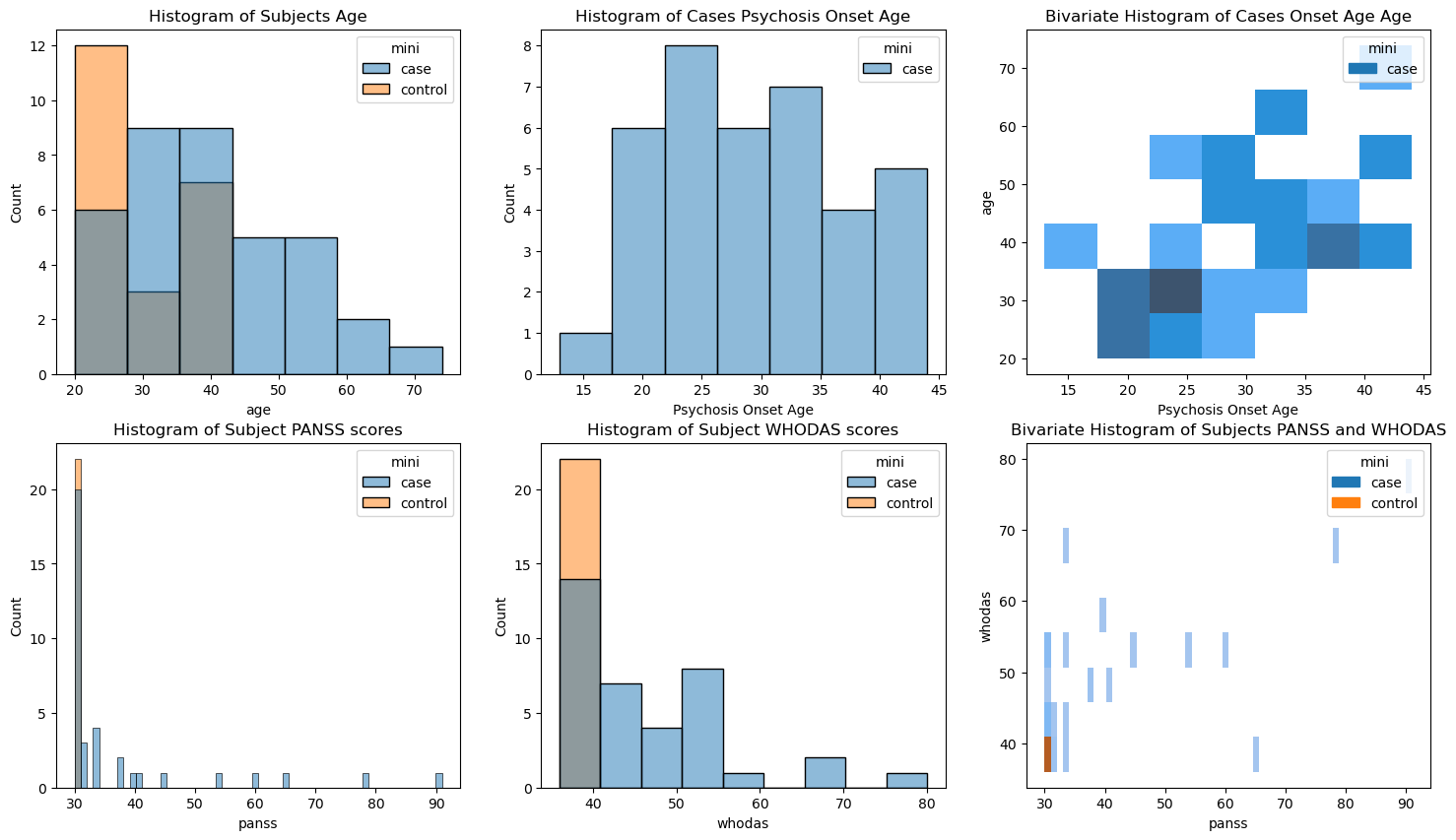}
    \caption{Demographic Information Plots}
    \label{fig:metric_plots}
    \hypertarget{fig:metric_plots}{}
\end{figure}

\section{Methods}
\subsection{Candidates Selection}
All EEG data acquisition participants were of legal age and provided explicit consent for their data’s public availability in schizophrenia studies. Before participation, candidates underwent rigorous diagnosis using Positive and Negative Symptoms Scale (PANSS), World Health Organization Disability Assessment Schedule (WHODAS), and the Mini International Neuropsychiatric Interview (MINI) criteria, categorizing them as healthy controls or patients based on stringent criteria  (see \hyperlink{fig:metric_plots}{Figure 2} for PANSS and WHODAS histograms). PANSS scores range from 30/210 for healthy controls, with higher scores for patients. WHODAS assesses functional disabilities, with healthy controls averaging 36/256, and patients scoring higher (refer to \hyperlink{fig:metric_plots}{Figure 2} for WHODAS score histogram).

Following nosological diagnosis, some candidates did not proceed to EEG data acquisition due to specific inclusion and exclusion criteria. For EEG inclusion, patients needed a schizophrenia diagnosis via the MINI interview, and those previously diagnosed needed active psychosis. Exclusion criteria included recent substance use (cannabis, hallucinogens, alcohol), bipolar disorder, or epilepsy. Legal age was required for consent, and family mental disorder history, though noted, was not a primary criterion. 

Selected candidates needed to be well-rested and abstain from alcohol, drugs, caffeine, or stimulants for at least three days before participating in a data-acquisition session. Those with hair conditions obstructing scalp access or who applied hair products within 24 hours before the EEG data-acquisition session were disqualified.

\subsection{Experiments, Task \& Protocol Design, Implementation}
Recognizing the consistent use of mismatch negativity, steady state response, and entropy measures in
schizophrenia diagnosis, and considering the condition’s heterogeneous nature, along with observed associations of auditory steady state responses with GABA (gamma-aminobutyric acid) neurotransmitter deficiencies in schizophrenia from literature, the robustness of entropy measures, and the efficacy of fuzzy entropy, the acquisition schemes which allow for multiple feature computation and thus study of different phenomenon in schizophrenia was adopted.

Existing EEG datasets for schizophrenia studies typically offer singular biomarkers, limiting research aiming to combine and evaluate multiple biomarkers. In NSzED data-acquisition sessions, each subject recorded a minimum of four and maximum of five EEG recordings. These recordings enable computation of event-related potentials, rhythmic oscillation parameters, and complexity measures individually or in combination, facilitating a more comprehensive analysis.

The EEG data-acquisition phases aimed to capture diverse neural and cognitive responses. Rest state recordings help establish a baseline for rest-state rhythmic oscillations. Arithmetic tasks during recording help to activate cognitive pathways, facilitating computation of complexity and organizational measures. Fixed frequency auditory stimuli at 40Hz during recording, were used with the aim of comparing rest-state rhythmic oscillation and auditory steady state response parameters. Auditory oddball paradigm experiments during recordings help elicit mismatch negativity and some other event-related potentials relevant to schizophrenia studies.

Time-locked labeling of stimulus presentation and task onset is achieved with the open-source lab streaming layer suite, synchronizing events with the EEG data stream. Information about each EEG device’s datasets, including phase duration, is stored in the respective dataset folders. Custom acquisition software allows for adjusting the recording duration for each phase.

\subsection{Participants}
Participants were recruited from the mental health ward and medical outpatient department of Obafemi Awolowo University Teaching Hospital College (OAUTHC) in Ile-Ife, Osun state, and the Wesley Guild Hospital in Ilesa, Osun state, within the age bracket of 20 to 70. Statistical description and comparison of patients and controls is shown in \hyperlink{tab:descriptive_statistics}{Tables 3-5}).

\begin{table}
  \centering
  \caption{Descriptive Statistics of patients and Controls}
  \small
  \begin{tabular}{p{1.5cm}cccccccccccccc}
    \hline
    {} & \multicolumn{7}{c}{Healthy Controls} & \multicolumn{7}{c}{Schizophrenia Patients} \\
    \addlinespace
    {} & Mean & SD 2 & Min 3 & 25\% 4 & 50\% 5 & 75\% & Max & Mean & SD & Min & 25\% & 50\% & 75\% & Max\\
    \hline
    Age & 28.95 & 7.65 & 20 & 23 & 25 & 36.75 & 42 & 40.22 & 12.75 & 20 & 30 & 39 & 47 & 74\\
    \addlinespace
    Psychosis Onset Age & - & -  & - & - & - & - & - & 24.93 & 7.93 & 13 & 23 & 29.34 & 37 & 44\\
    \addlinespace
    Duration of Illness & - & -  & - & - & - & - & - & 11.29 & 8.67 & 1 & 4 & 11 & 16 & 30\\
    \addlinespace
    No of Previous Admissions & - & -  & - & - & - & - & - & 1.24 & 1.42 & 0 & 0 & 1 & 2 & 7\\
    \addlinespace
    PANSS & 30 & 0 & 30 & 30 & 30 & 30 & 30 & 37.31 & 14.31 & 30 & 30 & 30 & 37.31 & 91\\
    \addlinespace
    WHODAS & 36 & 0 & 36 & 36 & 36 & 36 & 36 & 46.03 & 10.82 & 36 & 36 & 45 & 54 & 80\\
    \hline
  \end{tabular}
  \label{tab:descriptive_statistics}
  \hypertarget{tab:descriptive_statistics}{}
\end{table}

\begin{table}
  \centering
  \caption{Independent t-test Between patients and Controls Numerical Information.}
  \small
  \begin{tabular}{ccccccc}
    \hline
    {} & Age & Psychosis Onset Age & Duration of Illness & No of Previous Admissions & PANSS & WHODAS\\
    \hline
    \addlinespace
    t-value & 3.75 & 17.28 & 6.09 & 4.10 & 2.39 & 4.33\\
    \addlinespace
    p-value & 0.0004 & \textless 0.0001 & \textless 0.0001 & 0.0001 & 0.02 & \textless 0.0001\\
    \hline
  \end{tabular}
  \label{tab:inferential_statistics}
\end{table}

\begin{table}
  \centering
  \caption{Chi2 Test Between patients and Controls Nominal Inormation.}
  \small
  \begin{tabular}{ccccccc}
    \hline
    {} & Symptoms & Ethnicity & Education Level & Employment Status & Occupation & MINI\\
    \hline
    \addlinespace
    chi2 & 0 & 66 & 15.79 & 17.89 & 242 & 0\\
    \addlinespace
    p-value & \textless 0.0001 & \textless 0.0001 & \textless 0.0001 & 0.0001 & \textless 0.0001 & \textless 1\\
    \hline
  \end{tabular}
  \label{tab:inferential_chi2_statistics}
\end{table}

\subsection{Pre, During \& Post Data-Acquisition Precautions}
Before data acquisition, crucial preparatory measures were taken to ensure optimal EEG recordings. Initial steps included thorough subject preparation, including obtaining informed consent, explaining procedures, and ensuring subjects were well-rested and free from substances affecting brain activity. Subsequently, precise electrode placement using the standardized 10-20 system was meticulously done, with stringent checks on electrode impedance levels to keep impedance's below 5K\textOmega, minimizing potential noise interference in the recordings. 

Configuring EEG equipment involves meticulous calibration, determining optimal sampling rates for devices with configurable sampling rates, and setting filters for desired frequency ranges while reducing artifacts (50Hz Notch Filter). During EEG data-acquisition, preventive measures ensure data quality and participant safety, including maintaining low electrode impedance, monitoring and addressing artifacts, and verifying secure electrode attachment. Participants were instructed to minimize movements for reduced motion artifacts, prioritizing their comfort throughout the data acquisition session. Environmental controls minimized external interference, following safety protocols for potential emergencies involving seizures or aggressive psychotic episodes.

Following EEG data-acquisition sessions, clinical specialists meticulously analyze recorded EEG data. Their focus extends beyond identifying and mitigating noise or artifacts to detecting null activity, deviations from typical EEG patterns, and rhythmic patterns or seizure-related waveforms. Additionally, they monitor transitions between different states of brain activity, ensuring the presence of event-related potentials, and scrutinize patterns, trends, or fluctuations within the EEG data. Importantly, clinicians establish correlations between EEG recordings and subjects’ clinical history, symptoms, and diagnostic information, employing a comprehensive approach to accurately interpret the findings and inform clinical decision-making.

\subsection{Data Preprocessing}
After successful EEG data acquisition and a rigorous quality assurance review, preprocessing aimed
to reduce artifact, power-line interference noises and extract signals from frequency bands of interest (1-100 Hz). Processed EEG data for each subject was stored in designated folders for feature computation and model development. Unprocessed raw EEG data for each subject was also provided in the subject folder, allowing investigations into novel preprocessing techniques. All device data underwent baseline correction and were band-passed between 1-100 Hz band.

\section{Dataset Utility}
In an effort to demonstrate the utility of the acquired dataset, mismatch-negativity waveform and amplitudes, fuzzy entropy values and band specific steady-state amplitude and phase response were computed. The approach towards computing these features and results of basic analysis of the features are shown in this section.

\subsection{Mismatch Negativity (MMN)}
MMN, an event-related potential component in EEG responses to auditory stimuli, assesses the ability to detect deviant tones. MMN is associated with auditory processing deficits in schizophrenia. Standard stimuli include a 1KHz, 100ms tone. Additionally, a frequency deviant tone of 3KHz, 100ms, and a duration deviant of 1KHz, 200ms are employed.

The MMN waveforms are computed by averaging event-related potentials for each tone class separately for each subject within a 450ms window. The average of deviant tones is then subtracted from that of the standard tone to compute the MMN waveform. Traditionally, the peak amplitude in the 450ms window quantifies MMN, but this work instead used amplitudes from five windows: 0-100ms, 100-200ms, 200-300ms, 300-400ms, and 400-450ms, recognizing that latency for peak amplitudes could vary among subjects. \hyperlink{fig:avg_mmn_waveforms}{Figure 4} shows the average MMN waveforms of patients and controls while \hyperlink{fig:avg_mmn_values}{Figure 3} shows the montage of MMN amplitude values in five time windows computed from the MMN waveforms.

\begin{figure}
    \centering
    \includegraphics[width=1\textwidth]{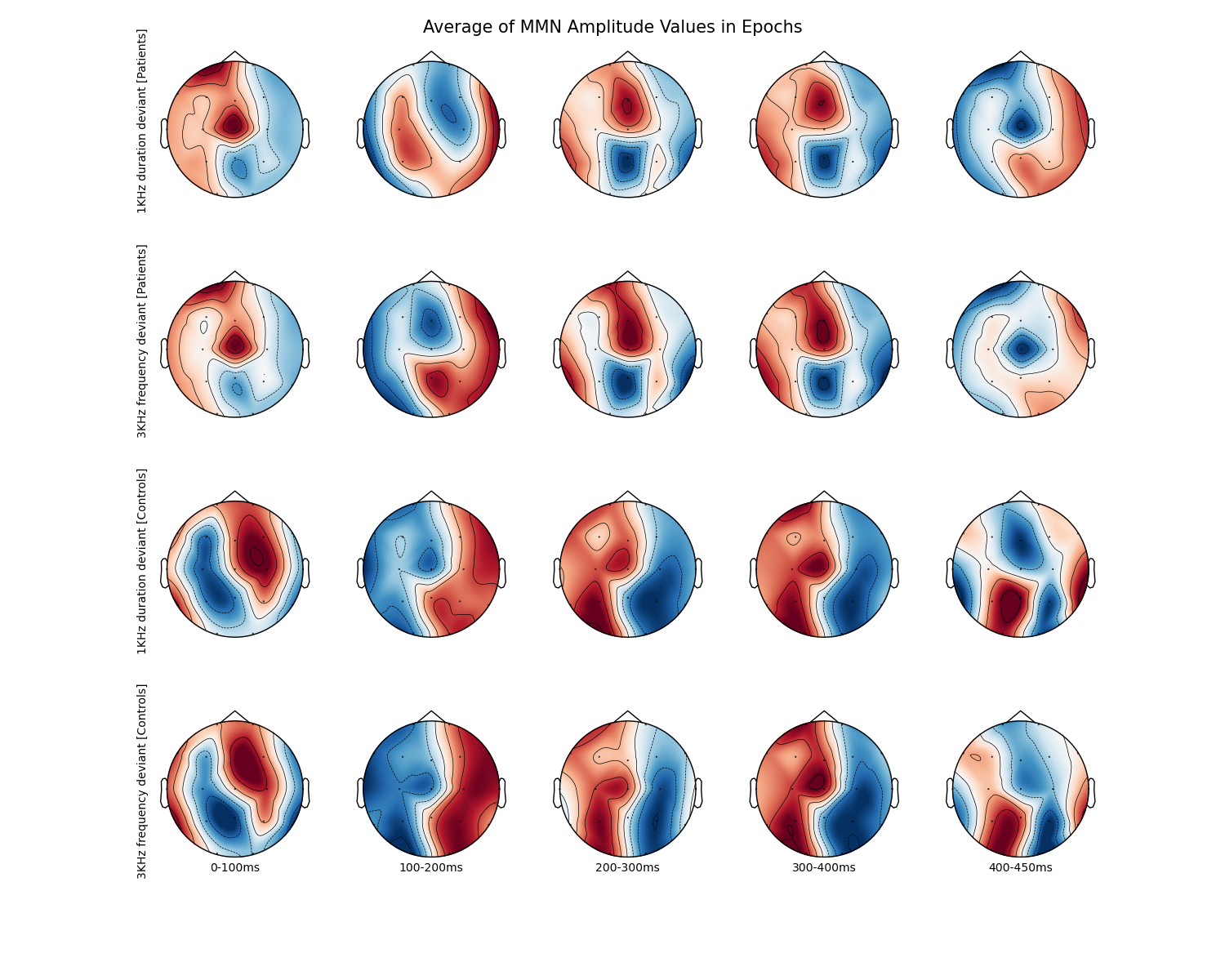}
    \caption{Average of Mismatch Negativity Amplitude Values Across Patients \& Controls}
    \label{fig:avg_mmn_values}
    \hypertarget{fig:avg_mmn_values}{}
\end{figure}
\begin{figure}
    \centering
    \includegraphics[width=1\textwidth]{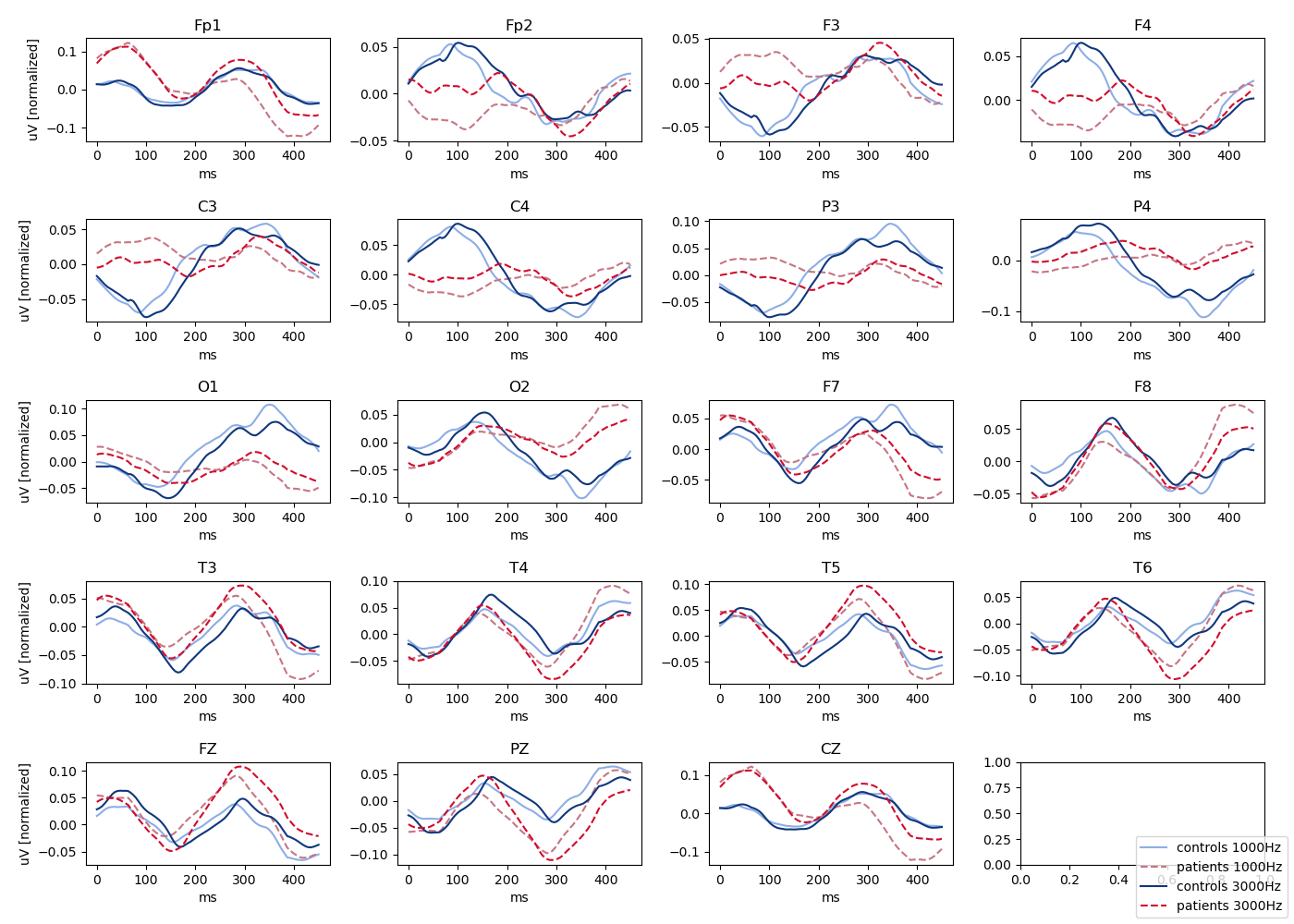}
    \caption{Average of Mismatch Negativity Waveforms Across Patients \& Controls}
    \label{fig:avg_mmn_waveforms}
    \hypertarget{fig:avg_mmn_waveforms}{}
\end{figure}

\subsection{Fuzzy Entropy}
Fuzzy entropy is a method employed to quantify uncertainty or randomness in datasets containing imprecise or fuzzy values, extending classical entropy which deals with precise values. In the context of classifying schizophrenia using EEG data, fuzzy entropy is used to assess the complexity and irregularity of brain activity, offering additional insights for classification models. Fuzzy entropy features are computed from data acquired during the arithmetic task phase and restful states, involving the calculation of cortical group activity by averaging signals from similar cortical regions. The mean of fuzzy entropy values from epochs for each electrode of a recording was used as the fuzzy entropy feature. Cortical regions considered include frontal, central, parietal, temporal, and occipital lobes. \hyperlink{fig:avg_ent_montage}{Figure 5} shows the montage of the average of fuzzy entropy values across patients and controls.

\begin{figure}
    \centering
    \includegraphics[width=1\textwidth]{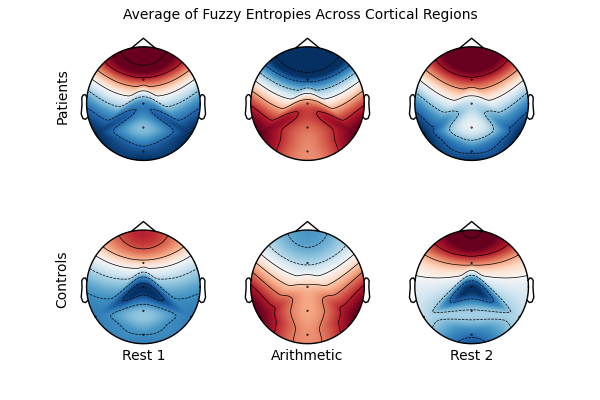}
    \caption{Average ofFuzzy Entropy Values Across Patients \& Controls}
    \label{fig:avg_ent_montage}
    \hypertarget{fig:avg_ent_montage}{}
\end{figure}

\subsection{Steady State Response}
Steady-state responses that occur in response to fixed stimuli or no stimuli also carry information, typically distinct in schizophrenia patients and healthy control populations. One such steady state response is the Auditory Steady State Response (ASSR) which represents EEG reactions to repetitive auditory stimuli. ASSR assesses the brain’s response to sound frequencies. During EEG acquisition, a 40Hz fixed-frequency tone elicits the ASSR. ASSR stimuli was presented only in recordings done using the BrainAtlas Discovery-24E device. Processing involves bandpass filtering at the tone frequency, fourier transform to get frequency representation, then the Hilbert transform for magnitude/phase information over time. A more intuitive alternative is computing the spectrogram to see synchronization to the tone frequency. As at the time of publication, computation of ASSR features was ongoing and as such no analysis is carried out on ASSR features yet.

\printbibliography

\end{document}